\documentclass{article}
\usepackage{spconf,amsmath,graphicx}
\usepackage{listings}
\usepackage{xcolor}
\lstnewenvironment{queryl}[1][] 
   {\lstset{frame=shadowbox,escapechar=`,linewidth=8cm, #1}}
   {}
\usepackage{enumitem}
\usepackage{multirow}
\usepackage{textcomp}
\usepackage[ruled,vlined]{algorithm2e}

\title{N-Best Hypotheses Reranking for Text-To-SQL Systems}
\name{Lu Zeng, Sree Hari Krishnan Parthasarathi, Dilek Hakkani-Tur}
\address{Alexa, Amazon, USA.\\
{\small \tt \{luzeng, sparta,hakkanit\}@amazon.com}}

\copyrightnotice{978-1-6654-7189-3/22/\$31.00~\copyright2023 IEEE}
\begin{document}
%
\maketitle
\begin{abstract}
Text-to-SQL task maps natural language utterances to structured queries that can be issued to a database. State-of-the-art (SOTA) systems rely on finetuning large, pre-trained language models in conjunction with constrained decoding applying a SQL parser. On the well established Spider dataset, we begin with Oracle studies: specifically, choosing an Oracle hypothesis from a SOTA model's 10-best list, yields a $7.7\%$ absolute improvement in both exact match (EM) and execution (EX) accuracy, showing significant potential improvements with reranking. Identifying coherence and correctness as reranking approaches, we design a model generating a query plan and propose a heuristic schema linking algorithm. Combining both approaches, with T5-Large, we obtain a consistent $1\% $ improvement in EM accuracy, and a $~2.5\%$ improvement in EX, establishing a new SOTA for this task. Our comprehensive error studies on DEV data show the underlying difficulty in making progress on this task.
\end{abstract}

%

\section{Introduction}
\label{sec:intro}
Large language models (LM) are widely used for natural language generation~\cite{2020t5, lewis2019bart, brown2020language}. Recently, the use of large LMs has extended to semantic parsing tasks such as code generation. For general-purpose code generation, large LMs such as Codex~\cite{chen2021evaluating} are trained on massive, paired codebases (code, NL). For domain-specific code generation tasks, such as Text-to-SQL that aims to convert natural language instructions to SQL queries, multiple public domain datasets are available (with associated leaderboards), including Spider~\cite{yu2018spider}, CoSQL~\cite{yu2019cosql} and SParC~\cite{yu2019sparc}. However, the amount of training data in these datasets is much smaller than the natural language and code pairs mined from the internet. In such cases, instead of training a new model from scratch, the pretrain/finetune strategy with publicly available LMs has been shown to be more accurate. For example, in UnifiedSKG~\cite{xie2022unifiedskg}, the T5 model~\cite{2020t5} is finetuned for various semantic parsing tasks (including Text-to-SQL), achieving SOTA performance. Besides the amount of the training data, another challenge with SQL generation is that the generated code is underspecified without the corresponding schema~\footnote{For Text-to-SQL, in addition to SQL grammar constraints, there are schema constraints.}. To handle this challenge, implicit or explicit schema linking becomes an important sub-task for SQL code generation~\cite{lei2020re,guo2019towards,yu2018typesql}.

In this paper, we focus on complex, cross-domain, SQL generation using Spider~\cite{yu2018spider} a large-scale, Text-to-SQL dataset consisting of 200 complex databases covering 138 domains. Spider is a well established dataset with nearly 70 entries on the leaderboard, demonstrating the difficulty of the task. The data is split into training, development (DEV), and test sets without overlaps in databases across these sets, as the aim of this task is to learn models that can issue queries from natural language text to previously unseen databases in the training set. Furthermore, SQL queries in this dataset contain nested sub-queries, requiring the model to understand compositional structures. The model performance is based on two metrics, exact-set-match accuracy (EM) and execution accuracy (EX); the former compares individual query components between the predicted and groundtruth SQL queries, while the latter compares their execution output.

Similar to general-purpose code generation, the output of Text-to-SQL models is constrained to follow SQL grammar. Previous solutions have employed encoder/decoder models, with the decoder being constrained to produce well-formed outputs~\cite{wang2019rat,yin2017syntactic}. An approach that is more compatible with large pretrained LMs is to remove the constraints on decoder:~\cite{suhr2020exploring, lin2020bridging} prune the finalized hypotheses from beam search to those that are syntactically correct. Meanwhile, PICARD~\cite{scholak2021picard}, a top entry on the Spider leaderboard~\footnote{https://yale-lily.github.io/spider}, in addition to finetuning the pretrained T5 model, also imposes SQL syntax constraints during beam search (via constrained decoding). It achieves an EM of 74.8\% and an EX of 79.2\% on the  Spider DEV set. Natural ways to potentially improve these metrics include increasing the model size or collecting and/or synthesizing additional training data~\cite{kaplan2020scaling}. 

In this paper, we take a different approach: we first perform an Oracle analysis on n-best lists obtained from PICARD and observe significant improvements (7.7\% absolute improvement in EM and EX) even at small beam sizes (such as 10). The gap in performance between 1-best and Oracle motivates reranking approaches. We propose 2 reranking approaches that are motivated by the issues with the current, large pretrained LMs: \textit{coherence}~\cite{nakano2021webgpt} and \textit{correctness}. To improve coherence, we explore n-best reranking using a query plan produced by an independent model. Next, to improve correctness, we propose a heuristic algorithm performing schema linking on n-best lists -- imposing constraints that are missing in PICARD. The combined reranking approaches consistently yield improvements across all T5 model sizes, obtaining a 1.0\% absolute improvement in EM and a 2.5\% absolute improvement on EX. Lastly, to analyze persistent gap with Oracle results, we performed a detailed analysis of errors and metrics. Our studies show that annotation and metrics issues can be significant factors to improving models and evaluation on Spider.

\noindent \textbf{Contributions.} The contributions of this paper are: a) Using Oracle analysis and rerankers on n-best lists, we outperform SOTA performance on a competitive Spider task; b) Analysis of errors on Spider DEV data, showing much work remains to be done on metrics and annotation.

\vspace{-2mm}
\section{Related Work}
\label{sec:relatedwork}
\vspace{-2mm}
We review related work in 4 categories: a) reranking approaches; b) coherence in LMs; c) schema linking for text-to-SQL; d) noise in Spider. 

\noindent \textbf{Rerankers.} N-best list reranking has a long history in speech recognition, machine translation, and semantic parsing communities~\cite{schwartz1996multiple, nguyen1997efficient, zens2005rwth, luong2014word,collins2005discriminative, ge2006discriminative, hui2021dynamic}. These are performed with a much larger $2^{nd}$-stage model~\cite{nogueira2020document,wang2021retrieval} or using additional knowledge sources not available during the first pass~\cite{rayner1994combining,jonson2006dialogue}. On Spider, a larger $2^{nd}$-stage reranker was successfully used in~\cite{kelkar2020bertrand}. Some reranking methods combine outputs from multiple systems~\cite{rosti2007combining} or use improved output confidence~\cite{bach2011goodness}. Imposing task specific constraints~\cite{suhr2020exploring, lin2020bridging} to remove invalid options can be a helpful strategy to improve reranking. In our work, in addition to the output target being structured queries (instead of natural language), the $1^{st}$-stage model is large (T5 family), and the baseline model imposes additional knowledge during beam search in the form of SQL parsing. Our proposed methods are designed for this setting.

\noindent \textbf{Coherence with query plan.} Coherence issues with LMs~\cite{porzel2003contextual} are associated with hallucination phenomenon~\cite{raunak2021curious,lee2018hallucinations}. While these are well-known problems in unstructured text generation with LMs~\cite{maynez2020faithfulness,holtzman2019curious,meister2021language,malkin2022coherence}, coherence issues with structured text generation is somewhat less unexplored:~\cite{li2020seqgensql} uses semantic coherence for data augmentation, while~\cite{yin2019reranking} uses a reconstruction model for reranking semantic parses, and ~\cite{suhr2020exploring, lin2020bridging,scholak2021picard} use parsers to improve coherence with SQL grammar. Our reranker is designed to improve semantic coherence, and it is distinct from~\cite{yin2019reranking} in that, our query plan model predicts the structure of a query (from natural language), and orders the n-best list for consistency.

\noindent \textbf{Correctness with schema linking.} Historically, schema linking (SL) has been explored for Text-to-SQL. Schema has been encoded in model input for small encoder/decoder models~\cite{lin2020bridging,bogin2019representing, chen2021shadowgnn} and large pretrained LMs~\cite{xie2022unifiedskg,scholak2021picard}. More work has been done on modeling schema linking~\cite{lei2020re}, where the role of SL in Text-to-SQL is recognized. RAT-SQL~\cite{wang2019rat}, a system that had been influential in text-to-SQL, proposed a framework to encode relational structure in the database schema and simultaneously address schema linking. In contrast, we use schema linking as a post-processor and a reranker on the output of large LMs.

\noindent \textbf{Noise in Spider.} Spider dataset~\cite{yu2018spider} has been well-explored in Text-to-SQL, and it's corpus noise are documented in related work~\cite{lei2020re,wang2019rat,choi2021ryansql}. One type of corpus noise comes from annotation errors in groundtruth SQL queries and typos in natural language queries~\cite{lei2020re}. Another source of the noise is in the pairing of only one groundtruth SQL query to each natural language query, which results in a large portion of predicted queries being incorrectly marked as ``mispredicted''~\cite{wang2019rat,choi2021ryansql}. 

\begin{figure}[t]
  \centering
  \includegraphics[width=1\linewidth]{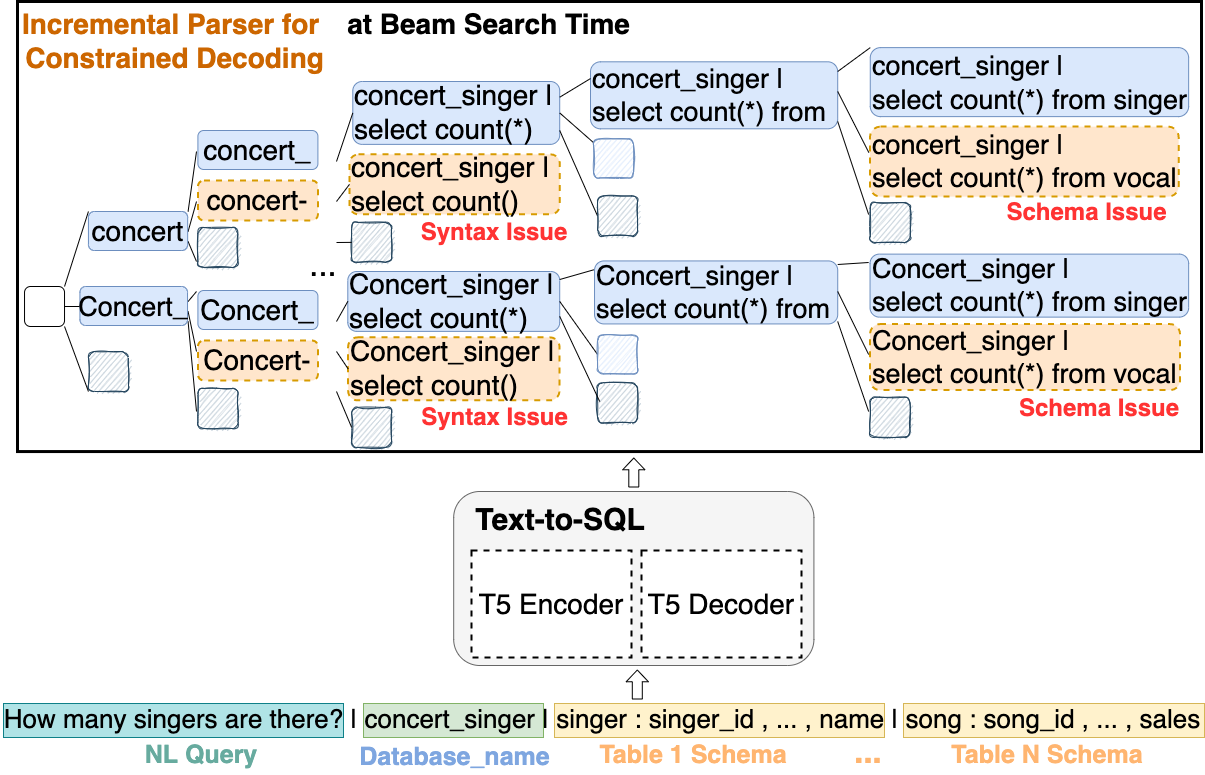}
  \vspace{-6mm}
  \caption{PICARD explained by an example. The prediction pattern is ``\texttt{\textlangle{}Database\_name\textrangle{} | \textlangle{}pred\_SQL\textrangle{}}''.}
  \label{fig:picard}
  \vspace{-3mm}
\end{figure}

\vspace{-2mm}
\section{Text-to-SQL Using Pre-trained LMs}
\label{sec:task}
\vspace{-2mm}
We use PICARD as our baseline system, and use it to produce the n-best hypotheses in this paper.
It uses finetuned T5 models, and implements incremental SQL parsing to constrain decoding during beam search. Input to the T5 model includes natural language query, database name, and database schema (\texttt{table\_name : col1 , ... , coln}). PICARD employs a post-processor during beam search, and to save computing time it only checks the next top-k tokens for validity. Figure~\ref{fig:picard} shows how predictions are generated when PICARD is enabled, using $k = 2$ and beam size of 3. Shadowed box means that the result is not expanded or explored for the next time stamp; blue (solid) boxes show valid SQL query prefixes, while orange (dashed) boxes show sequences that violate SQL syntax.

The incremental parser not only filters out hypotheses that violate the SQL grammar, but also imposes rudimentary schema checks. Fig.~\ref{fig:picard} gives two examples of filtering hypotheses, the first one is due to a syntax error (e.g., \texttt{count()}) and the other one is due to schema error (e.g., table name \texttt{vocal} does not exist in the database).

\section{Reranking Approaches}
\label{sec:method}
In this section, we discuss our two reranking approaches over n-best hypotheses.

\vspace{-2mm}
\subsection{Improving Coherence with Query Plan}
\label{sec:method_query_plan}
Text generation is a challenging topic, and large LMs tend to lose coherence~\cite{nakano2021webgpt, malkin2022coherence}. This is compounded in text-to-SQL with such models: apart from conforming to syntactic structures, models have to perform semantic mapping from natural language to SQL over long spans in complex questions involving compositionality, grouping/ordering/counting. Furthermore, nested queries are likely to appear with \texttt{WHERE, EXCEPT, UNION, and INTERSECT} clauses, while grouping/ordering/counting tend to occur with \texttt{GROUP BY, HAVING, ORDER BY, LIMIT}. 

We build a model that focuses specifically on this aspect: a multi-label classification model that generates a query plan predicting whether a SQL query contains the aforementioned 8 clauses. The output of this model is then used as a coherence check reranker against the n-best hypotheses from the baseline model. We finetune a RoBERTa-Large using the classification head with binary cross entropy (BCE) loss. The model is trained using the training partition of Spider dataset. Input of the model consists of a natural language query and a database schema, while the labels are obtained from groundtruth SQL queries.

\vspace{-2mm}
\subsection{Improving Correctness with Schema Linking}
\label{sec:method_value_linking}
The motivation for this approach comes from the fact that certain aspects of schema linking (values, columns, and tables) from natural language query (on the input side of the model) to a SQL query (on the output side of the model) can be hard for the model to learn to map: this is because these mappings tend to be irregular, and a relatively small training dataset may not cover a large fraction of the irregularities. For example, a primary key field on ``student identifcation'' could be named in several ways as a column such as ``stud\_id'', ``sid'', ``s\_id'' etc. Similarly, values that are strings can be arbitrarily represented, such as ``True'' vs ``T'' vs ``Yes''. 

We implement an algorithm to perform value linking in \texttt{WHERE} clauses. For each predicted SQL query in the n-best list, we follow three steps:
\begin{enumerate}[leftmargin=*]
\vspace{-3mm}
\item Extract slot names and their respective values from the conditions in the \texttt{WHERE} clause; then check if the slot value exists in any of the referenced tables in the \texttt{FROM} clause of the query; 
\vspace{-3mm}
\item  Obtain a list of candidate slot names and values, which are exact or partial occurrences of the column/table names and string values in the question with name-based and value-based linking described in RAT-SQL~\cite{wang2019rat};
\vspace{-3mm}
\item Perform prefix and abbreviation matches on slot values with categorical types in a table schema definition (such as matching ``left'' to ``L'').
\end{enumerate}

\section{Experiments}
\label{sec:exp_rst}
This section presents the experimental setup, baselines, Oracle analysis, and the main results.

\vspace{-3mm}
\subsection{Dataset and Metrics}
We briefly describe the dataset and metrics used in this paper. More details can be found in~\cite{yu2018spider}.

\vspace{-3.5mm}
\subsubsection{Dataset}
Spider is a large-scale, cross-domain paired text-to-SQL dataset: it contains 10,181 questions and 5,693 unique complex SQL queries on 200 databases (covering 138 domains), each with multiple tables. The standard protocol splits this into 8,659 examples on 146 databases for training, and 1034 examples on 20 databases are used for development (DEV);  the test set that includes 2,147 examples from 34 databases are held back. The split is done without overlaps of databases across these sets. 

\noindent \textbf{Task difficulty levels. }Spider categorizes the SQL query complexity into 4 difficulty levels: \textit{easy}, \textit{medium}, \textit{hard}, and \textit{extra hard}. Queries that contain more SQL keywords (\texttt{GROUP BY}, \texttt{ORDER BY}, \texttt{INTERSECT}), nested subqueries, column selections and aggregators are considered harder. 

\vspace{-3.5mm}
\subsubsection{Metrics}
On Spider, Text-to-SQL model performance is evaluated based on two metrics: exact-set-match accuracy (EM) and execution accuracy (EX). Note that all our results are on the Spider DEV set.

\noindent \textbf{Exact-set-match accuracy (EM).} EM compares each clause\footnote{Note that EM implemented by Spider ignores table join conditions.} between a prediction and its corresponding groundtruth SQL query. 
The predicted SQL query is correct only if all of the components match. This metric does not take values into account. EM can lead to false positives and false negatives.

\noindent \textbf{Execution Accuracy (EX).} EX compares the execution output of the predicted SQL query and its corresponding groundtruth SQL queries. However, it is important to note that EX can also lead to false positives and false negatives. 

\vspace{-3mm}
\subsection{Models}
\vspace{-1mm}

We use PICARD~\cite{scholak2021picard} as our baseline model in this paper. We finetune on three T5 model sizes, T5-Base\footnote{\label{note1}Note that the models are \textit{LM adapted}, initialized from T5 model and trained for additional 100K steps on the LM objective discussed in ~\cite{2020t5}}, T5-Large\footnotemark[4], and T5-3B on p3dn.24xlarge instances (8 NVIDIA Tesla V100 GPUs) and employ DeepSpeed to save memory\footnote{https://github.com/microsoft/DeepSpeed}. The input to the models contain several segments separated by \texttt{|}, including natural language query, database name, serialized database schema. We take the serialization scheme mentioned in~\cite{shaw2020compositional} and enable database content by appending database values to the column names~\cite{lin2020bridging}. All the T5 models are finetuned with teacher forcing and cross-entropy (CE) loss for up to 3000 epochs using a batch size of 2000 and a learning rate of $1e^{-4}$. During inference, an incremental SQL parser is integrated into beam search, with a beam size of 4. 

To improve coherence, we finetune a RoBERTa-Large model with a sequence classification head on p3.2xlarge instances (1 NVIDIA Tesla V100 GPU) for query plan prediction. We reuse the input from the baseline model. The output is one-hot encoding label extracted from groundtruth queries based on existence of \texttt{WHERE}, \texttt{GROUP BY}, \texttt{HAVING}, \texttt{ORDER BY}, \texttt{LIMIT}, \texttt{EXCEPT}, \texttt{UNION}, and \texttt{INTERSECT} clauses. The RoBERTa model is finetuned with binary cross entropy (BCE) loss for up to 100 epochs using a batch size of 5 and a learning rate of $1e^{-5}$.

\vspace{-3mm}
\subsection{Oracle Analysis}
\label{sec:oracle}
\vspace{-1mm}

We open the beam up and perform Oracle analysis on n-best hypotheses. Specifically, we conducted these experiments over a set of beam sizes (and correspondingly n-best hypotheses), 10, 15, 20, and 25; also a few selected T5 model sizes, T5-Base\footnotemark[4], T5-Large\footnotemark[4], and T5-3B. 

\begin{table}[h]
\vspace{-4mm}
\caption{1-best accuracies as beam size is changed.}
\vspace{1mm}
\label{table:picard_over_beam_size}
\centering
\begin{tabular}{|c|cc|cc|cc|}
\hline
{Beam} &  \multicolumn{2}{c|}{T5-Base} & \multicolumn{2}{c|}{T5-Large} & \multicolumn{2}{c|}{T5-3B}   \\
\multicolumn{1}{|c|}{\shortstack{size}}& EM\%   & EX\% & EM\%   & EX\% & EM\%   & EX\%\\
\hline
4        &   66.6 &   68.3 &     74.8 &   79.2&     75.5  & 79.3 \\
10       &   67.1 &   68.4 &     75.1 &   79.4&     75.6  & 79.3 \\
15       &   67.1 &   68.5  &    74.7&    79.2&     75.5  & 79.2 \\
20       &   67.3  &  68.7  &    74.8 &   79.2&     75.5  & 79.2 \\
25       &   67.3  & 68.7  &      74.7 &  79.1&     75.6  & 79.2 \\
\hline
\end{tabular}
\end{table}

\noindent \textbf{Search errors.} Table~\ref{table:picard_over_beam_size} presents the analysis by increasing the beam size (and choosing the 1-best hypothesis): opening up the beam, improves the accuracy by a small amount initially (going from 4 to 10) for all models. However, the performance on EM and EX saturate after that, suggesting that search errors contribute a small proportion of overall errors. 


\begin{table}[h]
\vspace{-2mm}
\caption{Oracle accuracies as beam size is changed.}
\vspace{1mm}
\centering
\label{table:oracle_over_beam_size}
\resizebox{\linewidth}{!}{
\begin{tabular}{|c|cc|cc|cc|l|}
\hline
{Beam} & \multicolumn{2}{c|}{T5-Base} & \multicolumn{2}{c|}{T5-Large} &\multicolumn{2}{c|}{T5-3B} &Note \\
\multicolumn{1}{|c|}{\shortstack{size}}  & EM\%   & EX\% & EM\%   & EX\% & EM\%   & EX\% & {}\\
\hline
10     &  67.1 &   68.3      &  75.1  &    79.4   & 75.6      & 79.3 & 1-best  \\
\hline
10      & 74.7      &  75.5     &   82.8 &    87.1  & 85.2      &   87.3  & \multirow{4}{*}{Oracle}\\
15      &  75.4     &  76.6     &   82.9 &    87.6  & 86.2      &   87.9 & \\
20      &  76.1     &  77.6    &   83.5 &    88.0  & 86.3    &   88.0 &  \\
25      & 76.5      &  78.3     &   83.8 &   88.0  & 86.8         &  88.8  & \\
\hline
\end{tabular}}
\end{table}

\noindent \textbf{Model errors.} In Table~\ref{table:oracle_over_beam_size}, the first row is the 1-best obtained from baseline, while the other rows show Oracle accuracies for the three models sizes over different beams. As can be observed, both EM and EX improve significantly: for example, the T5-Large model achieves 7.7\%, 7.8\%, 8.4\% and 8.7\% absolute improvements for EM; similarly 7.7\%, 8.2\%, 8.6\% and 8.6\% absolute improvements for EX. 

\begin{table}[!htbp]
\vspace{-3mm}
\caption{T5-large: 1-best and Oracle over difficulty levels.}
\vspace{1mm}
\centering
\label{table:t5_large_picard_oracle_over_diff}
\begin{tabular}{|c|c|cc|cc|cc|cc|}
\hline
Difficulty & count&  \multicolumn{2}{c|}{1-best} &  
\multicolumn{2}{c|}{Oracle}\\
{} & {}   & EM\%   &EX\%   & EM\%   &EX\%  \\
\hline
Easy        &  248  &  89.1  &   91.9   &   93.1    &  96.0 \\
Medium      &  446  &  80.9  &  84.8    &  88.6     &  92.6 \\
Hard        &  174  &  65.5  &  71.3    &  74.7     &  79.3  \\
Extra       &  166  &  48.8  &  54.8    &  60.2     &  67.5\\
Total       &  1034 &  75.1  &  79.4    &  82.8     &  87.1  \\
\hline
\end{tabular}
\vspace{-3mm}
\end{table}

\noindent \textbf{Difficulty levels.} Table~\ref{table:t5_large_picard_oracle_over_diff} compares the 1-best with Oracle using baseline (with T5-Large) over different difficulty levels. It can be observed that the Oracle results are consistently higher over both metrics (EM and EX) at all difficulty levels. The biggest improvement occurs on extra hard queries, with absolute improvements of more than 10\% for both metrics. Overall, these results suggest that significant improvements can be obtained using reranking approaches over n-best lists.

\vspace{-3mm}
\subsection{Results}
\label{sec:results}
\vspace{-1mm}

\subsubsection{Improving Coherence with Query Plan Modeling}
Table~\ref{table:qp_t5_large} lists reranking results using query plan (QP) coherence on the 10-best hypotheses obtained from the baseline using a T5-Large model. It can be seen that using structural consistency with QP can help both EM and EX, obtaining 0.7\% and 0.5\% absolute improvements respectively. We present an example improved by QP. A query plan that combines \texttt{GROUPBY} and \texttt{HAVING} clauses is presented below.

\begin{figure}[h!]
  \vspace{-2mm}
  \includegraphics[width=\linewidth]{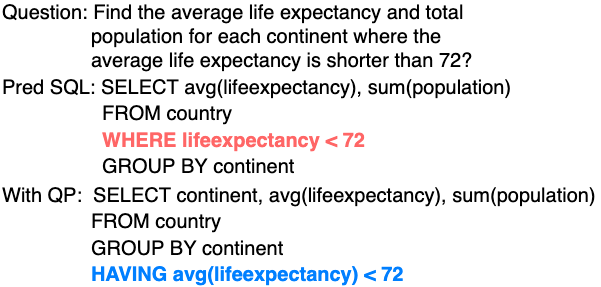}
\vspace{-7mm}
\end{figure}

\vspace{-4mm}
\begin{table}[h!]
\vspace{-2mm}
\caption{Reranking with query plan (QP) coherence checks on 10-best hypotheses from baseline using T5-Large.}
\vspace{1.5mm}
\centering
\label{table:qp_t5_large}
\begin{tabular}{|c|cc|cc|cc|}
\hline
Method &  \multicolumn{2}{c|}{T5-Large}   \\
{}   & EM\%   & EX\% \\
\hline
Baseline        &   75.1  & 79.4 \\
QP              &   75.8  & 79.9\\
\hline
\end{tabular}

\end{table}

To understand the QP model better, we present its performance on each of the clauses in Table~\ref{table:roberta_large_query_plan} in terms of F-1 score, precision and recall. While the model performs well on \texttt{WHERE, GROUP BY, HAVING, ORDER BY, LIMIT, INTERSECT}, it does not perform as well on \texttt{EXCEPT} and \texttt{UNION}. We conjecture low training data counts for these clauses as a potential reason. The macro F-1 score for this model is 0.89.

\begin{table}[h]
\vspace{-2mm}
\caption{QP model performance per clause.}
\vspace{1.5mm}
\centering
\label{table:roberta_large_query_plan}
\begin{tabular}{|c|c|c|c|}
\hline
\multicolumn{1}{|c|}{\shortstack{Clause}}	&	
\multicolumn{1}{c|}{\shortstack{F1}} &
\multicolumn{1}{c|}{\shortstack{R}} &
\multicolumn{1}{c|}{\shortstack{P}}  \\
\hline
 where      & 0.97  & 0.97  & 0.98 \\
groupBy     & 0.96  & 0.96  & 0.96 \\
having      &0.97   & 0.95  &0.99 \\
orderBy     &0.96   & 0.97  & 0.95 \\
limit       &0.95   & 0.96  & 0.94 \\
except      & 0.73  & 0.76  & 0.71 \\
union       &0.62   & 0.44  & 0.99  \\
intersect   & 0.96  & 0.97  &0.95 \\
\hline
\end{tabular}
\vspace{-2mm}
\vspace{-2mm}
\end{table}

\subsubsection{Improving Correctness with Schema Linking}
Table~\ref{table:schema_linking} shows the reranking performance with schema linking over 10-best hypotheses obtained from a baseline model with T5-Large. With this approach, we see an absolute improvement in EX of 1.9\%. 

\begin{table}[!htbp]
\vspace{-3mm}
\caption{Reranking with schema linking (SL) on 10-best hypotheses obtained from baseline using T5-Large.}
\vspace{1mm}
\centering
\label{table:schema_linking}
\begin{tabular}{|c|cc|}
\hline
Method &  \multicolumn{2}{c|}{T5-Large}  \\
{}     & EM\% &EX\%    \\
\hline
Baseline  & 75.1     & 79.4  \\
SL        & 75.4    &  81.3 \\
\hline
\end{tabular}
\end{table}
Below, we present 2 examples of schema linking errors (one of column name; and another of cell value) which are fixed by the proposed method. 

\begin{figure}[h!]
\vspace{-2mm}
\vspace{-2mm}
  \includegraphics[width=0.95\linewidth]{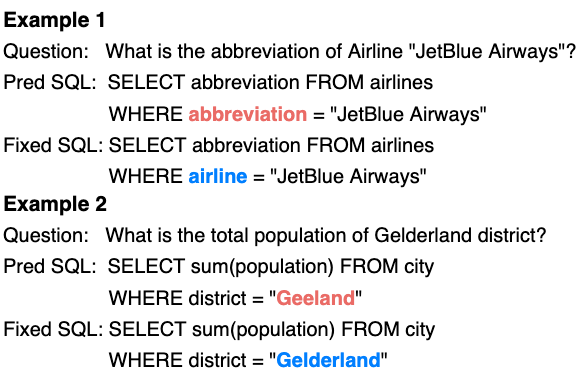}
\vspace{-2mm}
\end{figure}

\noindent In Ex 1, the question asks for the abbreviation of Airline ``JetBlue Airways'', where ``JetBlue Airways'' is a slot value for a slot name ``airline''. The model gets misled by an available column name that matches the query word ``abbreviation''. In Ex 2, we need a SQL query for the total population of ``Gelderland'' district. The expected slot value is ``Gelderland'', the predicted SQL obtained a cell value of ``Geeland''.

\vspace{2mm}
\noindent \textbf{Syntax Checks with a Full Parser.}
Since incremental parsing can still allow syntactically incorrect full hypotheses, we investigated the errors from the baseline model. Around 2\% of the hypotheses are indeed syntactically incorrect. For example, intermediate SQL queries from Fig.~\ref{fig:picard} such as \texttt{select count(*) from}, which are invalid queries but valid prefixes, are allowed by the incremental parser as outputs. We run a full SQL parser, filtering invalid hypotheses, and selecting the hypothesis with the highest remaining score.

\noindent \textit{Analysis.} 
Results from our experiments showed that the full parser does not improve performance. We investigated the errors for the 23 examples with syntactically incorrect 1-best predictions: for 16 of them all other hypotheses are incomplete SQL queries, while the other 7 do not have any hypotheses that improves either EM or EX. These are errors that cannot be improved even with an Oracle selection, meaning that these are residual search errors.
 
\vspace{-2mm}
\subsubsection{Combined Results}
We examine if the gains from SL and QP are additive, and if they carryover to different model sizes. We then combine SL and QP in order, presenting the results in Table~\ref{table:method_performance_of_methods_over_t5}. 

\begin{table}[h!]

\caption{Performance of rerankers (QP, SL, SL + QP) on 10-best lists obtained from different T5 model sizes.}
\vspace{1.5mm}
\centering
\label{table:method_performance_of_methods_over_t5}
\resizebox{\linewidth}{!}{
\begin{tabular}{|l|cc|cc|cc|}
\hline
Method &  \multicolumn{2}{c|}{T5-Base} & \multicolumn{2}{c|}{T5-Large} & \multicolumn{2}{c|}{T5-3B}  \\
{}   & EM\%   & EX\% & EM\%   & EX\% & EM\%  & EX\%\\
\hline
Baseline    &  67.1 & 68.3  & 75.1  & 79.4 & 75.6  & 79.3\\
\hline
QP         & 67.8  & 68.9  & 75.8  & 79.9  & 75.9  & 79.6 \\
SL         & 67.2  & 70.0  & 75.4  & 81.3  & 76.1  & 80.4\\
SL + QP      & 67.9  & 70.4  & 76.1  & 81.9  & 76.4  & 80.6\\
\hline
\end{tabular}}
\end{table}

From the table, it can be seen that the performance of the different rerankers are consistent across model sizes: the SL approach yields a consistent gain of between 1.1\% to 1.9\% in EX. Similarly, QP yields a consistent gain as well. Overall the gains combining the approaches are additive on both EM and EX across model sizes. Based on the one-sided t-tests, improvements on T5-base and T5-large are significant at $\alpha = 0.1$, while the gain on T5-3B is significant at $\alpha = 0.15$.

\vspace{-2mm}
\subsubsection{Analysis of gains across difficulty levels}
Table~\ref{table:gains_diff_levels} shows that the gains from the proposed approach (combining SL and QP) consistently carry over all the pre-defined difficulty levels. Furthermore, the gains are bigger on harder examples, considering both EM and EX.

\begin{table}[h!]
\vspace{-2mm}
\caption{T5-large model: Performance of rerankers (combining SL and QP) on 10-best lists}
\vspace{1mm}
\centering
\label{table:gains_diff_levels}
\begin{tabular}{|c|c|cc|cc|}
\hline
Diff & count&  \multicolumn{2}{c|}{Baseline} & \multicolumn{2}{c|}{SL + QP}\\
{} & {}   & EM\%   &EX\%   & EM\%      &EX\% \\
\hline
Easy   &  248 
        &  89.1   &   91.9 
        &   89.5  &    94.8 \\
Med   &  446  
        &  80.9  &  84.8  
        &  81.6  &  87.9  \\
Hard   &  174 
        &  65.5  &  71.3  
        &  68.4  &  73.6  \\
Extra   &  166  
        &  48.8  &  54.8  
        &  49.4 &  55.4  \\
\hline
Total   &  1034  
        &  75.1  &  79.4 
        &  76.1  &  81.9  \\
\hline
\end{tabular}
\end{table}

\section{Discussion on Errors}
\vspace{-2mm}
\label{sec:discussion_main}

Given that proposed reranker still has a gap with Oracle, we analyze errors with metrics computation and annotation.
\vspace{-1mm}
\subsection{Metrics Errors}

\noindent \textbf{PK/FK Swaps.} EM implementation by Spider is done by replacing foreign keys with primary keys. This is problematic and may lead to false positives. The following example contains a prediction that is marked as exact match incorrectly.  

\begin{figure}[h!]
\vspace{-1mm}
  \includegraphics[width=0.95\linewidth]{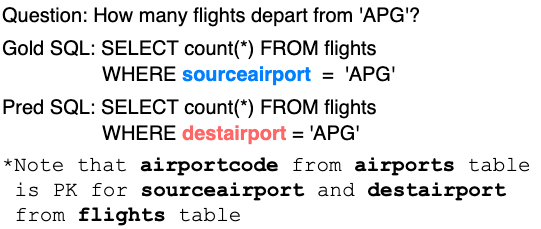}
  \vspace{-1mm}
\end{figure}

We fixed the EM implementation and obtained the revised $EM^*$ in Table~\ref{table:EM_revising}. Generally there is a 3-4\% absolute gap between the original and revised EM implementation.

\begin{table}[!htbp]
\vspace{-3mm}
\centering
\caption{Revised EM implementation.}
\vspace{1.5mm}
\label{table:EM_revising}
\begin{tabular}{|c|cc|cc|}
\hline
Method &  \multicolumn{2}{c|}{T5-large LM} &  \multicolumn{2}{c|}{T5-3b}  \\
{}   & $EM$\% &$EM^{*}$\%    & $EM$\%  &$EM^{*}$\% \\
\hline
PICARD    &  75.1  & 72.1&    75.6 &  71.9\\
SL + QP & 76.1  & 72.9   & 76.4   & 72.5\\
\hline
\end{tabular}
\vspace{-0mm}
\end{table}

\noindent \textbf{Query rewrite errors.}
Since SQL queries can sometimes be rewritten in multiple ways, EM can yield false negatives. We conducted error analysis over hypotheses selected after our combined reranking system (SL + QP) using finetuned T5-large model with PICARD. We found that 7.9\% of predictions are correct using EX, but incorrect with EM; also, 4.8\% of hypotheses are correct SQL but different to the corresponding groundtruth SQL. In such examples with SQL rewrites, EX serves as a complementary metric to EM. 

\vspace{2mm}
\noindent \textbf{Content-only match errors.}
As described in~\cite{yu2018spider}, it is possible to obtain a predicted SQL that returns the same result as the groundtruth SQL but semantically different. We found that 4.4\% of the hypotheses on DEV data generated from T5-Large model using PICARD are false positives of this type.

\vspace{2mm}
\noindent \textbf{Redundant column errors.}
We found that some predicted or groundtruth SQL queries contain redundant columns. The columns repeat information or do not need to be included, however retaining such column won't affect the semantics of the query, but can result in EX errors. 

\begin{figure}[h!]
  \vspace{-2mm}
  \includegraphics[width=\linewidth]{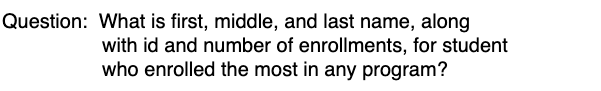}
 \vspace{-6mm}
\end{figure}

\begin{figure}[h!]
  \vspace{-3mm}
  \includegraphics[width=\linewidth]{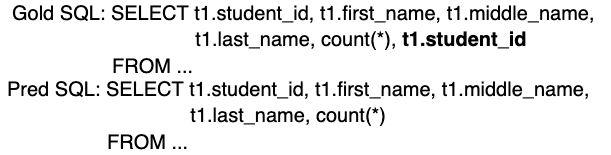}
 \vspace{-6mm}
\end{figure}

\vspace{-5mm}
\subsection{Annotation Errors}
We found 5 unique groundtruth SQL queries with annotation errors in DEV set. In Spider, some natural language queries are presented multiple times; consequently, the 5 unique groundtruth SQL queries are assigned to 7 DEV examples. We show an example with annotation error whose natural language query has been written in two different ways below. The annotation error is at least 0.7\%.

\begin{figure}[h!]
  \vspace{-1mm}
  \includegraphics[width=0.93\linewidth]{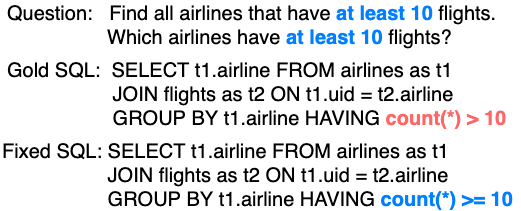}
  \vspace{-4mm}
  \vspace{-2mm}
\end{figure}

\vspace{-2mm}
\subsection{Summary and Outlook on Errors}
As we have seen, 0.7\% of the DEV examples have incorrect groundtruth SQL queries. We also identified issues with EM implementation in Spider -- foreign keys are automatically replaced with the corresponding primary keys, resulting in 3-4\% of the predictions on DEV examples being incorrectly marked as exact matches (false positives). Also, each example in Spider is annotated with only one groundtruth SQL -- leading to about 4.8\% of the predictions on DEV examples being false negatives. Using EX, results in 4.4\% of the predictions on DEV being false positive, while another 0.8\% of the predictions are false negative. Our studies show that metrics and annotation need further research attention on Spider.

\vspace{-1mm}
\section{Conclusions}
\label{sec:conclusions}
\vspace{-1mm}

We began with Oracle studies, showing large gains can be obtained with n-best hypotheses reranking for a SOTA Text-to-SQL system. Motivated by correctness and coherence issues in text generation using large PLMs, we proposed two reranking approaches: a) schema linking; b) query plan modeling. On a competitive Spider dataset, applying our proposed reranking methods to 10-best hypotheses obtained from a SOTA system (PICARD), we achieve improvements of $1\%$ in EM and $~2.5\%$ in EX. We analyzed errors in metrics computation and annotation -- showing these can be significant factors to improving models and evaluation on Spider.

\clearpage
\bibliographystyle{IEEEbib}
\bibliography{refs}

\begin{thebibliography}{10}

\bibitem{2020t5}
Colin Raffel, Noam Shazeer, Adam Roberts, Katherine Lee, Sharan Narang, Michael
  Matena, Yanqi Zhou, Wei Li, and Peter~J. Liu,
\newblock ``Exploring the limits of transfer learning with a unified
  text-to-text transformer,''
\newblock {\em Journal of Machine Learning Research}, vol. 21, no. 140, pp.
  1--67, 2020.

\bibitem{lewis2019bart}
Mike Lewis, Yinhan Liu, Naman Goyal, Marjan Ghazvininejad, Abdelrahman Mohamed,
  Omer Levy, Ves Stoyanov, and Luke Zettlemoyer,
\newblock ``Bart: Denoising sequence-to-sequence pre-training for natural
  language generation, translation, and comprehension,''
\newblock {\em arXiv preprint arXiv:1910.13461}, 2019.

\bibitem{brown2020language}
Tom Brown, Benjamin Mann, Nick Ryder, Melanie Subbiah, Jared~D Kaplan, Prafulla
  Dhariwal, Arvind Neelakantan, Pranav Shyam, Girish Sastry, Amanda Askell,
  et~al.,
\newblock ``Language models are few-shot learners,''
\newblock {\em Proc. of NeurIPS}, vol. 33, pp. 1877--1901, 2020.

\bibitem{chen2021evaluating}
Mark Chen, Jerry Tworek, Heewoo Jun, Qiming Yuan, Henrique Ponde de~Oliveira
  Pinto, Jared Kaplan, Harri Edwards, Yuri Burda, Nicholas Joseph, Greg
  Brockman, et~al.,
\newblock ``Evaluating large language models trained on code,''
\newblock {\em arXiv preprint arXiv:2107.03374}, 2021.

\bibitem{yu2018spider}
Tao Yu, Rui Zhang, Kai Yang, Michihiro Yasunaga, Dongxu Wang, Zifan Li, James
  Ma, Irene Li, Qingning Yao, Shanelle Roman, et~al.,
\newblock ``Spider: A large-scale human-labeled dataset for complex and
  cross-domain semantic parsing and text-to-sql task,''
\newblock {\em arXiv preprint arXiv:1809.08887}, 2018.

\bibitem{yu2019cosql}
Tao Yu, Rui Zhang, He~Yang Er, Suyi Li, Eric Xue, Bo~Pang, Xi~Victoria Lin,
  Yi~Chern Tan, Tianze Shi, Zihan Li, et~al.,
\newblock ``Cosql: A conversational text-to-sql challenge towards cross-domain
  natural language interfaces to databases,''
\newblock {\em arXiv preprint arXiv:1909.05378}, 2019.

\bibitem{yu2019sparc}
Tao Yu, Rui Zhang, Michihiro Yasunaga, Yi~Chern Tan, Xi~Victoria Lin, Suyi Li,
  Heyang Er, Irene Li, Bo~Pang, Tao Chen, et~al.,
\newblock ``Sparc: cross-domain semantic parsing in context,''
\newblock {\em arXiv preprint arXiv:1906.02285}, 2019.

\bibitem{xie2022unifiedskg}
Tianbao Xie, Chen~Henry Wu, Peng Shi, Ruiqi Zhong, Torsten Scholak, Michihiro
  Yasunaga, Chien-Sheng Wu, Ming Zhong, Pengcheng Yin, Sida~I Wang, et~al.,
\newblock ``Unifiedskg: Unifying and multi-tasking structured knowledge
  grounding with text-to-text language models,''
\newblock {\em arXiv preprint arXiv:2201.05966}, 2022.

\bibitem{lei2020re}
Wenqiang Lei, Weixin Wang, Zhixin Ma, Tian Gan, Wei Lu, Min-Yen Kan, and
  Tat-Seng Chua,
\newblock ``Re-examining the role of schema linking in text-to-sql,''
\newblock in {\em Proc. of EMNLP}, 2020, pp. 6943--6954.

\bibitem{guo2019towards}
Jiaqi Guo, Zecheng Zhan, Yan Gao, Yan Xiao, Jian-Guang Lou, Ting Liu, and
  Dongmei Zhang,
\newblock ``Towards complex text-to-sql in cross-domain database with
  intermediate representation,''
\newblock {\em arXiv preprint arXiv:1905.08205}, 2019.

\bibitem{yu2018typesql}
Tao Yu, Zifan Li, Zilin Zhang, Rui Zhang, and Dragomir Radev,
\newblock ``Typesql: Knowledge-based type-aware neural text-to-sql
  generation,''
\newblock {\em arXiv preprint arXiv:1804.09769}, 2018.

\bibitem{wang2019rat}
Bailin Wang, Richard Shin, Xiaodong Liu, Oleksandr Polozov, and Matthew
  Richardson,
\newblock ``Rat-sql: Relation-aware schema encoding and linking for text-to-sql
  parsers,''
\newblock {\em arXiv preprint arXiv:1911.04942}, 2019.

\bibitem{yin2017syntactic}
Pengcheng Yin and Graham Neubig,
\newblock ``A syntactic neural model for general-purpose code generation,''
\newblock {\em arXiv preprint arXiv:1704.01696}, 2017.

\bibitem{suhr2020exploring}
Alane~Laughlin Suhr, Kenton Lee, Ming-Wei Chang, and Pete Shaw,
\newblock ``Exploring unexplored generalization challenges for cross-database
  semantic parsing,''
\newblock 2020.

\bibitem{lin2020bridging}
Xi~Victoria Lin, Richard Socher, and Caiming Xiong,
\newblock ``Bridging textual and tabular data for cross-domain text-to-sql
  semantic parsing,''
\newblock {\em arXiv preprint arXiv:2012.12627}, 2020.

\bibitem{scholak2021picard}
Torsten Scholak, Nathan Schucher, and Dzmitry Bahdanau,
\newblock ``Picard: Parsing incrementally for constrained auto-regressive
  decoding from language models,''
\newblock {\em arXiv preprint arXiv:2109.05093}, 2021.

\bibitem{kaplan2020scaling}
Jared Kaplan, Sam McCandlish, Tom Henighan, Tom~B Brown, Benjamin Chess, Rewon
  Child, Scott Gray, Alec Radford, Jeffrey Wu, and Dario Amodei,
\newblock ``Scaling laws for neural language models,''
\newblock {\em arXiv preprint arXiv:2001.08361}, 2020.

\bibitem{nakano2021webgpt}
Reiichiro Nakano, Jacob Hilton, Suchir Balaji, Jeff Wu, Long Ouyang, Christina
  Kim, Christopher Hesse, Shantanu Jain, Vineet Kosaraju, William Saunders,
  et~al.,
\newblock ``Webgpt: Browser-assisted question-answering with human feedback,''
\newblock {\em arXiv preprint arXiv:2112.09332}, 2021.

\bibitem{schwartz1996multiple}
Richard Schwartz, Long Nguyen, and John Makhoul,
\newblock ``Multiple-pass search strategies,''
\newblock in {\em Automatic Speech and Speaker Recognition}, pp. 429--456.
  Springer, 1996.

\bibitem{nguyen1997efficient}
Long Nguyen and Richard~M Schwartz,
\newblock ``Efficient 2-pass n-best decoder.,''
\newblock in {\em EuroSpeech}. Citeseer, 1997.

\bibitem{zens2005rwth}
Richard Zens, Oliver Bender, Sa{\v{s}}a Hasan, Shahram Khadivi, Evgeny Matusov,
  Jia Xu, Yuqi Zhang, and Hermann Ney,
\newblock ``The rwth phrase-based statistical machine translation system,''
\newblock in {\em Proc. of IWSLT}, 2005.

\bibitem{luong2014word}
Ngoc-Quang Luong, Laurent Besacier, and Benjamin Lecouteux,
\newblock ``Word confidence estimation for smt n-best list re-ranking,''
\newblock in {\em Proc. of the Workshop on HaCaT during EACL}, 2014.

\bibitem{collins2005discriminative}
Michael Collins and Terry Koo,
\newblock ``Discriminative reranking for natural language parsing,''
\newblock {\em Computational Linguistics}, vol. 31, no. 1, pp. 25--70, 2005.

\bibitem{ge2006discriminative}
Ruifang Ge and Raymond Mooney,
\newblock ``Discriminative reranking for semantic parsing,''
\newblock in {\em Proc. of COLING/ACL}, 2006, pp. 263--270.

\bibitem{hui2021dynamic}
Binyuan Hui, Ruiying Geng, Qiyu Ren, Binhua Li, Yongbin Li, Jian Sun, Fei
  Huang, Luo Si, Pengfei Zhu, and Xiaodan Zhu,
\newblock ``Dynamic hybrid relation exploration network for cross-domain
  context-dependent semantic parsing,''
\newblock in {\em Proceedings of the AAAI Conference on Artificial
  Intelligence}, 2021, vol.~35, pp. 13116--13124.

\bibitem{nogueira2020document}
Rodrigo Nogueira, Zhiying Jiang, and Jimmy Lin,
\newblock ``Document ranking with a pretrained sequence-to-sequence model,''
\newblock {\em arXiv preprint arXiv:2003.06713}, 2020.

\bibitem{wang2021retrieval}
Zhiguo Wang, Patrick Ng, Ramesh Nallapati, and Bing Xiang,
\newblock ``Retrieval, re-ranking and multi-task learning for knowledge-base
  question answering,''
\newblock in {\em Proc. of the European Chapter of ACL: Main Volume}, 2021, pp.
  347--357.

\bibitem{rayner1994combining}
Manny Rayner, David Carter, Vassilios Digalakis, and Patti Price,
\newblock ``Combining knowledge sources to reorder n-best speech hypothesis
  lists,''
\newblock {\em arXiv preprint cmp-lg/9407010}, 1994.

\bibitem{jonson2006dialogue}
Rebecca Jonson,
\newblock ``Dialogue context-based re-ranking of asr hypotheses,''
\newblock in {\em SLT}. IEEE, 2006, pp. 174--177.

\bibitem{kelkar2020bertrand}
Amol Kelkar, Rohan Relan, Vaishali Bhardwaj, Saurabh Vaichal, Chandra Khatri,
  and Peter Relan,
\newblock ``Bertrand-dr: Improving text-to-sql using a discriminative
  re-ranker,''
\newblock {\em arXiv preprint arXiv:2002.00557}, 2020.

\bibitem{rosti2007combining}
Antti-Veikko Rosti, Necip~Fazil Ayan, Bing Xiang, Spyros Matsoukas, Richard
  Schwartz, and Bonnie Dorr,
\newblock ``Combining outputs from multiple machine translation systems,''
\newblock in {\em HLT 2007: The Conference of NAACL; Proc. of the Main
  Conference}, 2007, pp. 228--235.

\bibitem{bach2011goodness}
Nguyen Bach, Fei Huang, and Yaser Al-Onaizan,
\newblock ``Goodness: A method for measuring machine translation confidence,''
\newblock in {\em Proc. of the 49th Annual Meeting of ACL: HLT}, 2011, pp.
  211--219.

\bibitem{porzel2003contextual}
Robert Porzel and Iryna Gurevych,
\newblock ``Contextual coherence in natural language processing,''
\newblock in {\em International and Interdisciplinary Conference on Modeling
  and Using Context}. Springer, 2003, pp. 272--285.

\bibitem{raunak2021curious}
Vikas Raunak, Arul Menezes, and Marcin Junczys-Dowmunt,
\newblock ``The curious case of hallucinations in neural machine translation,''
\newblock {\em arXiv preprint arXiv:2104.06683}, 2021.

\bibitem{lee2018hallucinations}
Katherine Lee, Orhan Firat, Ashish Agarwal, Clara Fannjiang, and David
  Sussillo,
\newblock ``Hallucinations in neural machine translation,''
\newblock 2018.

\bibitem{maynez2020faithfulness}
Joshua Maynez, Shashi Narayan, Bernd Bohnet, and Ryan McDonald,
\newblock ``On faithfulness and factuality in abstractive summarization,''
\newblock {\em arXiv preprint arXiv:2005.00661}, 2020.

\bibitem{holtzman2019curious}
Ari Holtzman, Jan Buys, Li~Du, Maxwell Forbes, and Yejin Choi,
\newblock ``The curious case of neural text degeneration,''
\newblock {\em arXiv preprint arXiv:1904.09751}, 2019.

\bibitem{meister2021language}
Clara Meister and Ryan Cotterell,
\newblock ``Language model evaluation beyond perplexity,''
\newblock {\em arXiv preprint arXiv:2106.00085}, 2021.

\bibitem{malkin2022coherence}
Nikolay Malkin, Zhen Wang, and Nebojsa Jojic,
\newblock ``Coherence boosting: When your pretrained language model is not
  paying enough attention,''
\newblock in {\em Proc. of the 60th Annual Meeting of ACL (Volume 1: Long
  Papers)}, 2022, pp. 8214--8236.

\bibitem{li2020seqgensql}
Ning Li, Bethany Keller, Mark Butler, and Daniel Cer,
\newblock ``Seqgensql--a robust sequence generation model for structured query
  language,''
\newblock {\em arXiv preprint arXiv:2011.03836}, 2020.

\bibitem{yin2019reranking}
Pengcheng Yin and Graham Neubig,
\newblock ``Reranking for neural semantic parsing,''
\newblock in {\em Proc. of the 57th Annual Meeting of ACL}, 2019.

\bibitem{bogin2019representing}
Ben Bogin, Matt Gardner, and Jonathan Berant,
\newblock ``Representing schema structure with graph neural networks for
  text-to-sql parsing,''
\newblock {\em arXiv preprint arXiv:1905.06241}, 2019.

\bibitem{chen2021shadowgnn}
Zhi Chen, Lu~Chen, Yanbin Zhao, Ruisheng Cao, Zihan Xu, Su~Zhu, and Kai Yu,
\newblock ``Shadowgnn: Graph projection neural network for text-to-sql
  parser,''
\newblock {\em arXiv preprint arXiv:2104.04689}, 2021.

\bibitem{choi2021ryansql}
DongHyun Choi, Myeong~Cheol Shin, EungGyun Kim, and Dong~Ryeol Shin,
\newblock ``Ryansql: Recursively applying sketch-based slot fillings for
  complex text-to-sql in cross-domain databases,''
\newblock {\em Computational Linguistics}, vol. 47, no. 2, pp. 309--332, 2021.

\bibitem{shaw2020compositional}
Peter Shaw, Ming-Wei Chang, Panupong Pasupat, and Kristina Toutanova,
\newblock ``Compositional generalization and natural language variation: Can a
  semantic parsing approach handle both?,''
\newblock {\em arXiv preprint arXiv:2010.12725}, 2020.

\end{thebibliography}

\end{document}